# Wheelchair Behavior Recognition for Visualizing Sidewalk Accessibility by Deep Neural Networks

Takumi Watanabe[1][0000-0003-2432-239X], Hiroki Takahashi[1], Goh Sato[1], Yusuke Iwasawa[2], Yutaka Matsuo[2], and Ikuko Eguchi Yairi[1]

[1] Graduate School of Science and Engineering, Sophia University, 7-1 Kioi-cho, Chiyoda-ku, Tokyo 102-8554, Japan
[2] Graduate School of Technology Management for Innovation, The University of Tokyo, 7-3-1 Hongo, Bunkyo-ku, Tokyo 113-8656, Japan
watanabe@yairilab.net

**Abstract.** This paper introduces our methodology to estimate sidewalk accessibilities from wheelchair behavior via a triaxial accelerometer in a smartphone installed under a wheelchair seat. Our method recognizes sidewalk accessibilities from environmental factors, e.g. gradient, curbs, and gaps, which influence wheelchair bodies and become a burden for people with mobility difficulties. This paper developed and evaluated a prototype system that visualizes sidewalk accessibility information by extracting knowledge from wheelchair acceleration using deep neural networks. Firstly, we created a supervised convolutional neural network model to classify road surface conditions using wheelchair acceleration data. Secondly, we applied a weakly supervised method to extract representations of road surface conditions without manual annotations. Finally, we developed a self-supervised variational autoencoder to assess sidewalk barriers for wheelchair users. The results show that the proposed method estimates sidewalk accessibilities from wheelchair accelerations and extracts knowledge of accessibilities by weakly supervised and self-supervised approaches.

## 1 Introduction

Providing accessibility information on sidewalks for people with mobility difficulties, such as older, mobility-impaired, and visually impaired people, is an important social issue. One solution to this issue using information and communication technology is to develop an accessibility map as a large geographic information system (GIS) to provide accessibility information [Laakso *et al.*, 2011; Karimi *et al.*, 2014]. In the existing methods for gathering large-scale accessibility information, experts evaluate sidewalk accessibilities from their images [Ponsard and Vincent, 2006], or accessibility information is recruited from volunteers by crowdsourcing [Hara, 2014]. These methods depend on human labor and are impractical when collecting accessibility information in a huge area. The recent expansion of intelligent gadgets, such as smartphones and smartwatches, familiarizes people with sensing their activities [Swan 2014]. Focusing on the fact that the acceleration signals of wheelchairs are influenced by a road surface condition, we have been proposing a system that evaluates sidewalk accessibilities from wheelchair accelerometer using machine learning. Notably, a wheelchair body is influenced by road surface conditions, e.g., gradient, curbs, and gaps, which become a burden for people with mobility difficulties. Human activities measured by body-worn sensors are recognized by applying machine learning [Wang *et al.*, 2019]. The possibility of various machine learning methods is investigated for activity recognition using mobile sensors [Plötz *et al.*, 2011]. Aiming at improving the recognition performance, convolutional neural networks (CNN) [Yang *et al.*, 2015], recurrent neural networks [Edel and Enrico, 2016], and their hybrid model [Yao *et al.*, 2017] are investigated.

In this paper, we introduce our methodology to estimate sidewalk accessibilities by recognizing road surface conditions from wheelchair acceleration signals. Our goal is to realize a system that provides services to visualize sidewalk accessibilities and navigate safely designed routes for users. We developed and evaluated a CNN model to classify road surface conditions, a weakly supervised model to extract highly representative knowledge from acceleration signals without human annotations, and a self-supervised autoencoder model to assess the degree of sidewalk barriers for wheelchair users and visualize accessibility information.

## 2 Sidewalk Accessibility Visualization

### 2.1 Proposed System

This section introduces our proposed system for providing accessibility information that is helpful for all pedestrians, especially people with difficulties with moving. Figure 1 shows an overview of the system. The wheelchair sensor signals are measured by a sensing application downloaded on the user's mobile device or installed in the wheelchair. The wheelchair acceleration database is created by measured signals and annotations. After training deep neural networks, knowledge of road conditions is extracted from the trained network. Then sidewalk accessibility information is accumulated as a sidewalk feature dataset and visualized as a navigation map.

The simplest type of accessibility visualization using human sensing is simple wheelchair trails [Mora *et al.*, 2017]. Wheelchair trails provide practical information for wheelchair users regarding wheelchair accessible roads and facilities. Although the information is useful, it is not sufficient for all wheelchair users. The trail approach indicates if someone could travel in a location, but wheelchair users may have different mobility and accessibility requirements. The physical abilities of wheelchair users are more diverse than generally imagined; some users are trained like Paralympic athletes, whereas others may damage their bodies with only a few vibrations. Critical information for wheelchair users includes the physical state of the road surface, such as the angle of a slope, the height of a curb, and the roughness of a road surface. This information about the physical state of the road surface helps all people with mobility difficulties as well as wheelchair users to make decisions about access/avoidance of a road according to their physical conditions and abilities. Therefore, the information about the physical state of the road is the foundation of road accessibility.

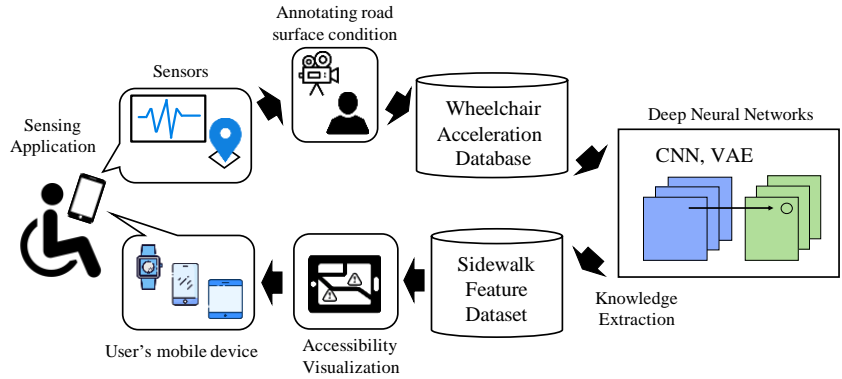

**Fig. 1.** Overview of the proposed system. Sidewalk accessibilities are visualized as a navigation map by using wheelchair sensor data and deep neural networks.

The vibrations of a wheelchair body are used for estimating the road accessibility information because wheelchair tires are directly influenced by the state of the road surface. Useful information is extracted from noisy raw signals of accelerometers installed in various wheelchairs because of the recent success of developing impersonal models by deep neural networks [Wang *et al.*, 2019]. Since extracting the only influence of road surface conditions from raw acceleration signals is challenging [Lara and Labrador, 2012], the observed wheelchair acceleration signals must be converted into an index that represents the road surface condition. Our ultimate goal is to realize a system that provides road accessibility visualization services to every user by using impersonal models that improve its accuracy as new data is provided by users. As the wheelchair traveling data in more diverse places are gathered by more users after the service is launched, the model is incrementally strengthened. Along with the maturity of the model, it will also be possible to extract the road accessibility information from the running data of baby strollers and bicycles as from wheelchairs. This paper aims to establish a fundamental method of knowledge extraction from wheelchair behavior data using deep neural networks in supervised classification and weakly supervised or self-supervised representation learning.

### 2.2 Related Work

Various mobility support systems for people with mobility difficulties have been proposed. GIS applications are utilized to create a walking space network composed of information about width, step, gradient, and its location of the walking route [Yairi and Igi, 2007; Zimmermann-Janschitz, 2018]. A navigation system for wheelchair users is provided on users' smartphones using on-site surveys [Koga *et al.*, 2015]. Although these mobility support systems are useful for impaired people, these systems depend on human labor, and collecting accessibility information in a huge area is impractical.

The quality of the road including depressions is detected based on a triaxial accelerometer and a gyroscope in android application using machine learning [Allouch *et al.,* 2017]. Abnormal traffic conditions in cities are detected using multimodal sensors of smartphones [Mohan *et al.*, 2008; Yu *et al.*, 2016]. Although these methods evaluate road conditions using automatic processing, collecting detailed road surface conditions, such as gradient, curbs, and the roughness of a road surface, remains difficult. Therefore, we focused on developing a system to extract accessibility knowledge from wheelchair acceleration signals that can be automatically collected and are influenced by road surface conditions.

In the existing machine learning methods, a large dataset with teacher labels is required to learn road surface conditions from acceleration data. Manual annotation of the dataset depends on human labor, which is both expensive and impractical to collect extensively. The weakly supervised learning [Zhou, 2018] method and the unsupervised feature learning [Längkvist *et al.*, 2014] method have been applied to various machine learning tasks, including human activity recognition [Sargano *et al.*, 2017], to avoid human annotations. This paper introduces our methodology to use weakly supervised and self-supervised approaches to extract accessibility knowledge by learning road surface conditions without human annotations from wheelchair acceleration signals.

## 3    Estimate Sidewalk Accessibilities

### 3.1    Dataset

The actual wheelchair driving data were collected to evaluate the proposed method. A total of nine wheelchair users between 20 and 60 years of age, including six manual wheelchairs and three electric wheelchairs, participated in the experiment. Their behaviors while driving about 1.4 km specified route (shown in Figure 2) around Yotsuya station in Tokyo were measured by a triaxial accelerometer in the iPod touch installed under a wheelchair seat, and positional data were measured using the quasi-zenith satellite system (QZSS). Acceleration values of the x, y, and z axes of the accelerometer were sampled at 50 Hz, and a total of 1,341,602 samples (about 7.5 hours) were collected. To confirm the circumstances when the acceleration data were measured, a video was recorded for both participants' driving state and the road surface conditions. Most of the entire route was a standard sidewalk, and a part of the course was a crosswalk. This route was carefully designed to include various road surface conditions to evaluate the generalization performance of the proposed method for common roads. If a user did not experience problems when moving up and down wheelchair ramp slopes, an excessive burden on the body and risk of an accident were considered minimal. Each participant drove the route in three laps; they drove clockwise from the start point to the goal point for the first and third laps and drove counterclockwise for the second lap. The slope and the gentle slope that were ascending clockwise in Figure 2 were ascending for the first and third laps and descending for the second lap.

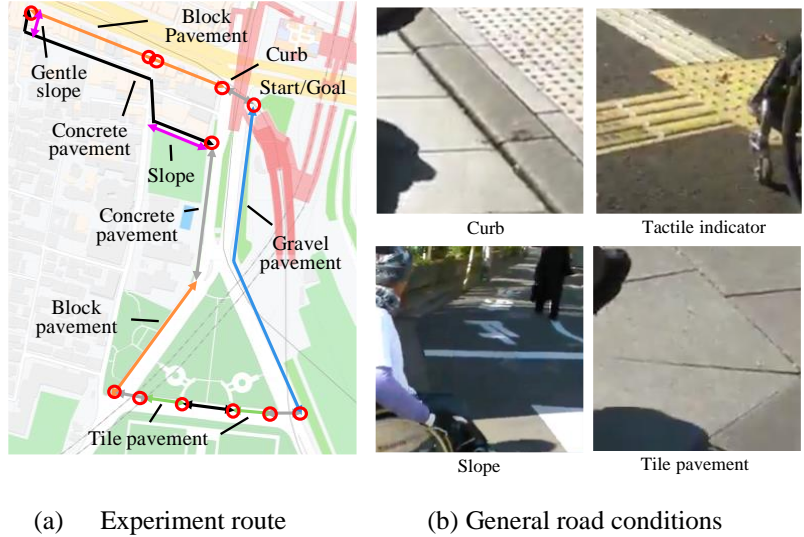

(a)   Experiment route  (b)   General road conditions

**Fig. 2.** Wheelchair experiment route. (a) The experimental route driven by wheelchair users. (b) The general road conditions in the route.

A mean filter with a length of five was processed for the acceleration dataset to remove noises. Then, the acceleration dataset was normalized to have a zero mean and unit standard deviation in each axis. To input acceleration data to a CNN model, the acceleration data were segmented into 29,727 examples using a sliding window. The window size was fixed to 450 (about nine seconds) with 90% overlap. The window size and the overlap percentage were selected to be adapted for the dataset following the procedure [Iwasawa *et al.,* 2016], which applied machine learning for wheelchair acceleration data.

### 3.2   Classifying Road Surface Conditions

This section provides a supervised CNN model to classify road surface conditions using wheelchair acceleration data. Road surface condition labels have four classes: moving on slopes (*Slope*), climbing on curbs (*Curb*), moving on tactile indicators (*TI*), and others (*Oths*). Each category represents typical road surface conditions: a continuous gradient, an abrupt step, a continued unevenness, and other conditions, respectively. These labels were created by visually observing the participants and the road surface conditions over the whole experiment video. The information on these road conditions directly conveys accessibility information to people with mobility difficulties, especially wheelchair users.

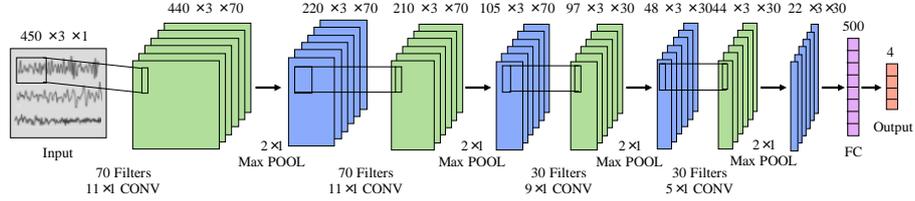

**Fig. 3.** The structure of the convolutional neural network to classify road surface conditions from wheelchair triaxial accelerations.

**Model Architecture.** Figure 3 shows the structure of CNN used to classify wheelchair acceleration data. The network is composed of an input layer, four convolution layers, one fully connected layer, and an output layer. The convolution layer consists of a convolution, a ReLU activation, and max-pooling processing. The fully connected layer consists of 500 units and a ReLU activation. The output layer is governed by a Softmax function that has four classes. The network has dropout layers after every convolution layer and fully connected layer. The dropout percentage is set to 20%, 30%, 30%, 40%, and 50% from the top to the bottom layer. This network follows the relevant research for recognizing human activity data [Yang *et al.*, 2015; Takahashi *et al.*, 2018], and the hyperparameters are adjusted for recognizing wheelchair accelerations. The Adam algorithm was used as an optimizer, and the learning rate was set to 0.0001. The test dataset was created by using a leave-one-subject-out (LOSO) methodology that verifies the performance for unknown users by evaluating a model with a dataset of a user who was not included in the training data. In this study, the model was trained repeatedly with a dataset of eight users as a training dataset, and the trained model was tested with the dataset of the remaining one user. The model score was evaluated by the mean of a total of nine trials. The validation dataset was created by dividing the training dataset into 90% training and 10% validation data using stratified splitting. The network was trained until the categorical cross-entropy loss of validation data stopped decreasing.

**Classification Result.** Table 1 compares the classification performance among the existing machine learning methods and the CNN model. Macro averaged F1 score (FS) and accuracy (Acc) were used as the evaluation index. Since *Oths* represented nearly 77% of the dataset, the classes are imbalanced, and the high F-score means that a model well recognizes barriers with few spots appearing in the dataset.

The first comparison method (*Raw + k-NN*) used raw acceleration signals as a feature set and used a k-nearest neighbor (k-NN) as a classifier. The second method (*MV + k-NN*) used the mean and the standard deviation (MV) of each axis of each segment as a feature set and used k-NN as a classifier. The third and fourth methods used rich heuristic features as a feature set. The following 12 types of values of each axis of each segment were computed as heuristic features: mean; standard deviation; maximum; minimum; zero-crossing; mean, standard deviation, maximum, and minimum of the difference; FFT frequency component; energy and entropy of the FFT frequency component. The heuristic features were classified by SVM with rbf kernel (*Heuristic + SVM*) and multilayer perceptron with two 500 units fully connected layers (*Heuristic + MLP*). The parameter $k$ of k-NN and the regularization parameter $C$ and the kernel coefficient parameter $\gamma$ of SVM were chosen using five-fold cross-validation to maximize the macro F-score. The parameter $C$ of SVM and the loss function of MLP were adjusted to inversely proportionally weight to class frequencies in the training data to handle the class imbalance problem. To ensure the fairness of the comparison between these methods and CNN, the activation of trained CNN was classified by SVM (*CNN + SVM*) and MLP (*CNN + MLP*). Motivated by previous research [Cao *et al.*, 2012], a smoothing method was implemented to post-process the predicted labels to enhance the prediction performance of the classifiers. Since the adjacent road surface conditions are in a similar state, the sample labels have a smooth trend. This smoothing method employs a low-pass filter to remove the impulse noise and maintain the edges. The impulse noise is a potential incorrect prediction, and the edges, in this case, are the transition in the road surface conditions. For the $i$th example, a smoothing filter with length seven was applied on the sequence whose center was the $i$th example. The predicted probabilities of the sequence were averaged for each class, and the class with the highest probability was assigned to the $i$th example.

Table 1. Comparison between hand-crafted feature classification methods and CNN model in F-score (FS) and accuracy (Acc).

| Method | FS | Acc (%) | FS | Acc (%) |
|---|---|---|---|---|
| | Without smoothing | | With smoothing | |
| Raw + k-NN | 28.2 | 75.5 | 26.4 | 74.9 |
| MV + k-NN | 45.0 | 69.9 | 45.6 | 73.2 |
| Heuristic + SVM | 51.9 | 80.3 | 51.8 | 80.7 |
| Heuristic + MLP | 56.5 | 78.2 | 56.3 | 78.9 |
| CNN + SVM | 62.6 | 82.5 | 67.4 | 85.2 |
| CNN + MLP | **68.7** | **84.7** | **71.3** | **86.4** |

The CNN method using raw wheelchair triaxial acceleration signals to classify road surface conditions achieved higher classification scores than the existing machine learning methods. This result shows that the proposed method is reasonable and practical to estimate road accessibilities.

### 3.3 Weakly Supervised Knowledge Extraction

#### 3.3.1 Methodology

This section introduces our weakly supervised method to extract representations of road surface conditions [Watanabe *et al.*, 2020]. Our method uses positional information collected while driving as low-cost weak supervision to learn road surface conditions and does not depend on human annotations. The positional information can be automatically collected with acceleration signals and notably is semantically related to road surface conditions. In the task of weakly supervised feature learning, determining what information to use for supervision is an important factor that affects the learning performance. We attempted to use a novel method incorporating positional information during wheelchair driving as weak supervision. Since adjacent road surfaces have similar conditions, our model effectively learns road surface conditions by being trained to predict the measured position of the input acceleration data. For the positional information in this paper, we confirmed the position where the acceleration data were measured by visually observing the experiment video to correct errors included in the QZSS positional data.

**Procedure to Generate Weak Supervision.** As the first step, the earth's surface was divided into a mesh shape. The objectives of dividing the earth's surface are to aggregate adjacent road surfaces into one group and to create discrete classes to formulate the position prediction task as a classification problem. The width of each grid created by the mesh was selected to 5 m in both the vertical and horizontal dimensions because a grid width under 5 m can distinguish sidewalks on both sides of a road with two or more lanes. Then, only the grids that covered the driving route were used as target grids. Finally, a unique number was assigned to each grid of the target grids. These assigned numbers are the identification (ID) of each grid. These IDs were assigned to all acceleration samples. The grid to which each sample belongs was identified by its positional data. These assigned IDs are the positional label set and are used as weak supervision for acceleration data. Through these steps, the same class is assigned to the adjacent road surface, and the CNN model is considered to effectively learn feature representations of road surface conditions.

**Position Prediction.** The model architecture and the training procedure of weakly supervised CNN follow those of supervised CNN in Section 3.2, except that weakly supervised CNN used positional labels and the output classes were the IDs of the grids. We hereafter refer to this CNN model trained on the weakly supervised task as the PosNet model.

The result of the position prediction task of PosNet was compared to that using another machine learning method. The accuracy of the PosNet model was 11.2%. As a comparison of the proposed model, FFT frequency components of each axis of each example were calculated and classified by logistic regression. The regularization parameter $C$ was chosen using five-fold cross-validation to maximize accuracy. The accuracy of the logistic regression was 5.86%. The pure chance was 0.32% in this case. The accuracy of these models is the mean of the total of nine trials obtained using the LOSO methodology. Although the score is low, the purpose of training the PosNet model with weak supervision was to learn feature representations of road surface conditions and accumulate them in the network. This absolute score is not important for evaluating the model.

### 3.3.2 Analysis

This section evaluates the representations of road surface conditions learned by the PosNet model. To evaluate the learned representations, the PosNet model was trained with the training dataset. Then the test dataset was input to the trained network, and the activation of the fully connected layer was obtained. This activation is a set of feature vectors and is the internal representation learned by the model from the input acceleration data. The obtained feature vector set was grouped by the k-means clustering to evaluate how well the learned representation conveys the road surface condition information. The clustering results were color-coded for each cluster and were plotted on the position of the input acceleration example on a map.

The route consisted of following 11 general road surface conditions: gravel pavement (GRAV), tile pavement (TILE), block pavement-1 (BLK-1), block pavement-2 (BLK-2), concrete pavement-1 (CONC-1), concrete pavement-2 (CONC-2), curb (CURB), ascending slope (ASC-SLP), descending slope (DESC-SLP), gentle ascending slope (GENT-ASC-SLP), and gentle descending slope (GENT-DESC-SLP). The clustering results were evaluated in detail by visually observing their plots for each lap. Figure 4 shows the visualization of the clustering result of one manual wheelchair user when the number of clusters $k$ was 16. The ASC-SLP of both the first and third laps were clearly grouped into the purple cluster, and GENT-ASC-SLP in both the first and third laps were grouped into the light purple cluster. DESC-SLP and part of GENT-DESC-SLP for the second lap were grouped into the light green cluster. The most CURB were grouped into the yellow-green cluster on every lap. For other pavement types, GRAV and CONC-1 were grouped into the blue cluster for every lap, and most parts of CONC-2b were grouped into the green cluster for every lap. Although the clustering result of GRAV and CONC were similar for every lap, the clustering tendency of TILE, BLK-1, and BLK-2 were different depending on the driving direction. This overall clustering tendency was observed in all nine dataset patterns.

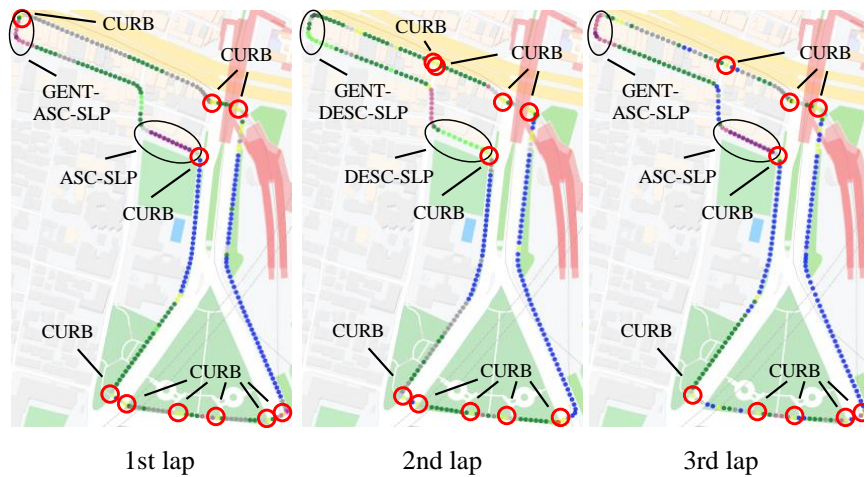

**Fig. 4.** The clustering visualization of the result of one manual wheelchair user.

These observations demonstrate that the PosNet model learns feature representations of detailed road surface conditions. ASC-SLP and DESC-SLP were grouped into separate independent clusters, although the same position label was assigned to the adjacent road regardless of the driving direction. This result shows that the model effectively learned representations of the differences between the ascending and the descending gradient. ASC-SLP and GENT-ASC-SLP were grouped into separate clusters. This result shows that the model learned representations of slight differences in gradient. The exact points of most CURB were grouped into the same cluster for any lap. This result shows that the model learned representations of wheelchair driving patterns over curbs. As a summary of the clustering results of three laps, all 12 curb points were detected. The same pavement types were roughly grouped into the same cluster, and the different pavement tended to be grouped into different clusters.

The PosNet model was found to learn rich feature representations of road surface conditions through the clustering evaluation. The usefulness of the learned representation was evaluated for recognizing general road surface conditions. The classification task of road surface conditions was evaluated in a semi-supervised setting. A common scenario for a semi-supervised setting is that a large amount of data is available and only a small fraction is labeled. This scenario is realistically expected for wheelchair data because acceleration data and positional information of wheelchairs can be extensively collected, and manual annotation to all acceleration data is expensive and impractical. Since the positional information can be automatically collected, the PosNet model was trained with the entire dataset. Then a classifier was trained with a subset of road surface condition labels and their corresponding feature set, which was obtained from the trained PosNet model. *Heuristic + MLP* and *CNN + MLP* were selected for comparison to the proposed method (*PosNet + MLP*). In the case of *CNN + MLP*, CNN was trained only with the subset of the training data because the CNN model was trained with road surface condition labels.

Figure 5 shows the transition of the classification performance under the semi-supervised setting. The 100% subset is the extreme case of using the entire dataset. The proposed model (*PosNet + MLP*) exceeds the performance of the fully supervised method (*CNN + MLP*) when the amount of labeled data decreases below 10%. The performance gap between them increased as the amount of labeled data decreased. The proposed method (*PosNet + MLP*) always outperformed *Heuri + MLP* on any subset proportion. This result demonstrates the usefulness of the proposed method in a practical environment. When more extensive wheelchair driving data are collected than the experiment conducted in this paper, the performance of the proposed method improves even if the amount of labeled data is limited, providing a highly practical model.

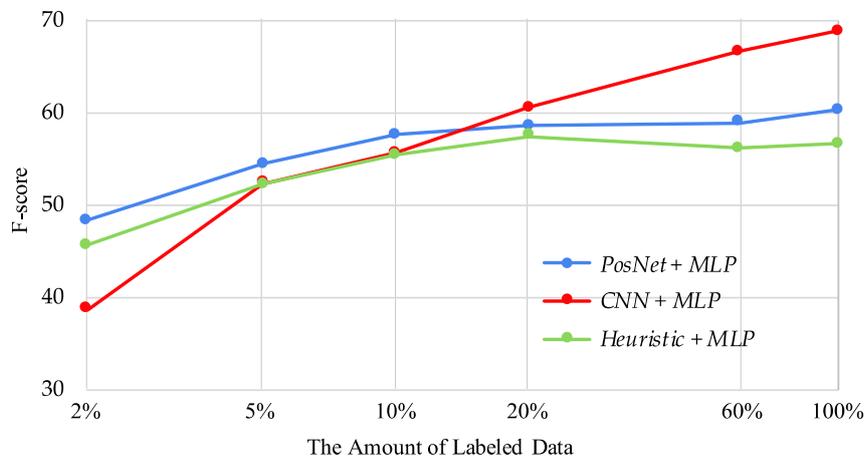

**Fig. 5.** The classification performance in a semi-supervised setting. The classification was implemented under 100%, 60%, 20%, 10%, 5%, and 2% subsets. The x-axis is logarithmic scale percentage of the amount of data with road surface condition labels. The y-axis is the macro F-score of the four classes.

### 3.4 Self-supervised Knowledge Extraction

#### 3.4.1 Methodology

This section introduces our self-supervised method to assess the degree of sidewalk barriers for wheelchair users. This method attempts to use convolutional variational autoencoder (ConvVAE) to extract knowledge of sidewalk accessibilities from wheelchair acceleration signals. ConvVAE can detect anomalies even when no or small labeled training data is provided. The reconstruction error calculated by ConvVAE is examined whether it reflects the road conditions and is used as the degree of burdens for wheelchair users.

The ConvVAE network is composed of encoder and decoder. The encoder has an input layer, four convolution layers, and the decoder has four deconvolution layers and a linear output layer. The latent distribution of the encoder is sampled to latent representation that has standard distribution and is decoded to reconstruct input acceleration through the decoder. The convolution layer in the encoder consists of a convolution, a ReLU activation, and max-pooling processing, and the deconvolution layer in the decoder consists of a deconvolution, a ReLU activation, and up-sampling processing. The kernel size and the number of feature maps of convolution layers and training procedure follow those of supervised CNN in Section 3.2, and the mean square error was used as a loss function of ConvVAE. The window size in segmenting acceleration signals was fixed to 400 in this experiment so that the latent representation is reconstructed to the same shape of input. The reconstruction error was calculated using the mean square error between the input and output signals.

### 3.4.2 Analysis

The reconstruction error calculated by the proposed ConvVAE is plotted on the corresponding position on a map and examined whether it reflects the degree of burdens for wheelchair users. After training the ConvVAE model by training dataset, the test dataset was input to the trained network, and the reconstruction error was obtained for each nine users by LOSO. The reconstruction error was normalized for each user and the maximum reconstruction error value over nine users were selected within every 5 m on the route.

Figure 6 shows the visualization of the maximum reconstruction error over nine users. The color of the plot points is determined by the ratio of each selected reconstruction error value to the largest value. The larger the ratio, the closer to red, and the smaller the ratio, the closer to blue. The large reconstruction errors are found at five CURB points and Unusual spot-A in the figure. Since the spot-A was under construction when the experiment was conducted, the participant drove around the spot differently from other roads. In most parts of the route, the values of reconstruction error were moderate or low. This result shows that the reconstruction error reflects the unusualness of the road such as abrupt steps and spots in an unusual environment. Since these unusual spots become burdens for wheelchair users, the ConvVAE model is indicated to extract some accessibility knowledge from wheelchair acceleration signals without any supervision.

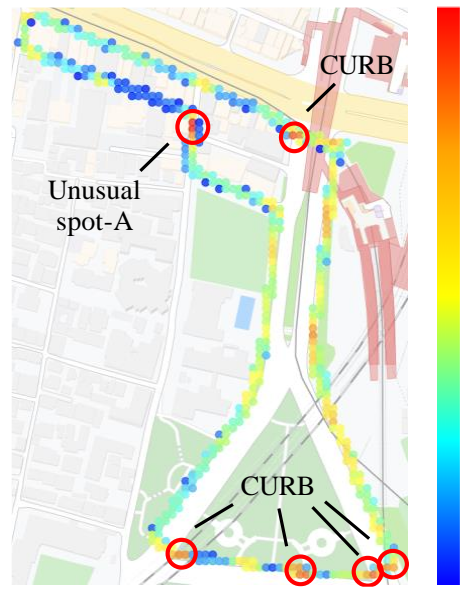

**Fig. 6.** The reconstruction error plot over nine participants.

## 4 Conclusion

The contributions of this paper were to confirm the possibility of using wheelchair acceleration signals to provide accessibility information. This paper developed and proposed a prototype system for visualizing sidewalk accessibility information that helps pedestrians, especially people with mobility difficulties. The proposed methodology used deep neural networks to estimate sidewalk accessibility by extracting knowledge from wheelchair behavior via a triaxial accelerometer in a smartphone installed under a wheelchair seat. The supervised method demonstrated that CNN classifies road surface conditions by recognizing wheelchair behavior from acceleration signals. A novel method was proposed to estimate road surface conditions without manual annotation by applying weakly supervised learning. The proposed method demonstrated that positional information during wheelchair driving helps with learning rich representations of road surface conditions, and the learned representations were highly discriminative for a road surface classification task. The learned representations were visualized on a map and demonstrated to provide detailed representations of road surface conditions, such as the difference of ascending and descending of a slope, the angle of slopes, the exact locations of curbs, and the slight differences of similar pavements. The learned representations were found to be more useful than calculated rich heuristic features for the road surface classification task. The self-supervised section introduced a method to assess the degree of sidewalk barriers for wheelchair users. The proposed method attempted to use ConvVAE to extract knowledge of sidewalk accessibilities from wheelchair acceleration signals. The result indicated to extract accessibility knowledge, such as abrupt steps and a spot in an unusual environment that leads to a barrier for wheelchair users, without any supervision.

Our future work will be directed to the improvement of the wheelchair behavior recognition model and the design of supervision that do not require human labor and are beneficial to estimate sidewalk accessibilities. Employing recent hybrid deep models [Yao *et al.*, 2017] are prospective to enhance the recognition model. The design of supervision includes unsupervised representation learning that could be employed for activity recognition [Ji *et al.*, 2019].

**Acknowledgments**

We would like to show our best gratitude to all participants in data collection. This research was supported by a Grant-in-Aid for Scientific Research (B), 17H01946 and 20H04476, Japan Society for the Promotion of Science.